\documentclass[conference]{IEEEtran}
\IEEEoverridecommandlockouts
\usepackage{cite}

\ifCLASSINFOpdf
  \usepackage[pdftex]{graphicx}
\else
  \usepackage[dvips]{graphicx}
\fi
\usepackage{atbegshi}
\usepackage{amsmath}
\usepackage{amssymb}
\usepackage{mathtools}
\usepackage{algorithmic}
\AtBeginShipout{%
  \ifnum\value{page}=1
    \def\PageDrop{17.5pt}%
  \else
    \def\PageDrop{4pt}%
  \fi
  \setbox\AtBeginShipoutBox=\vbox to \ht\AtBeginShipoutBox{%
    \vskip\PageDrop
    \hbox{%
      \kern7.15pt%
      \pdfliteral page {q 0.9775 0 0 1 0 0 cm}%
      \box\AtBeginShipoutBox
      \pdfliteral page {Q}%
    }%
    \vss
  }%
}

\newif\ifreview
\reviewfalse      

\usepackage{siunitx}
\usepackage{xcolor}
\usepackage[normalem]{ulem}
\usepackage{bm}

\newif\ifshowrevisions
 \showrevisionsfalse

\newcommand{\add}[1]{\ifshowrevisions\textcolor{blue}{#1}\else#1\fi}
\newcommand{\delete}[1]{\ifshowrevisions\textcolor{red}{\sout{#1}}\else\fi}
\newcommand{\revise}[2]{\ifshowrevisions\delete{#1}\add{#2}\else#2\fi}
\newcommand{\resistancename}{resistance} 

\begin{document}
%
\title{From Impact to Insight: Dynamics-Aware Proprioceptive Terrain Sensing on Granular Media}
%
%
%

\ifreview
  \author{Author Names Omitted for Anonymous Review}
\else
  \author{
  Yifeng~Zhang$^{1}$,
  Yue~Wu$^{1}$,
  Jake~Futterman$^{1}$,
  Jacob~Meseha$^{1}$,\\
  Eduardo~Rosales$^{1}$,
  Irie~Cooper$^{1}$,
  J.~Diego~Caporale$^{1,*}$,
  and Feifei~Qian$^{1,*}$%
  \thanks{$^{1}$All authors are with the Department of Electrical and Computer Engineering,
  University of Southern California, Los Angeles, CA 90089, USA.}%
  \thanks{$^{*}$Corresponding authors: J.~Diego~Caporale and Feifei~Qian.
  Emails: {\tt\small \{caporale, feifeiqi\}@usc.edu}.}%
  \thanks{First author email: {\tt\small yifengz@usc.edu}.}%
  }
\fi

%
%

\markboth{Journal of \LaTeX\ Class Files,~Vol.~14, No.~8, August~2015}%
{Shell \MakeLowercase{\textit{et al.}}: Bare Demo of IEEEtran.cls for IEEE Journals}
%



\maketitle

\begin{abstract}
Robots that traverse natural terrain must interpret contact forces generated under highly dynamic conditions.
However, most terrain characterization approaches rely on quasi-static assumptions that neglect velocity- and acceleration-dependent effects arising during impact and rapid stance transitions.
In this work, we investigate granular terrain interaction during high-speed hopping and develop a physics-based framework for dynamic terrain characterization using proprioceptive sensing alone.
Through controlled hopping experiments with systematically varied impact speed and leg compliance, our measurements reveal that quasi-static based assumptions lead to large discrepancies in granular terrain property estimation during high-speed hopping, particularly upon touchdown and controller-induced stiffness transitions.
Velocity-dependent drag alone cannot explain these discrepancies.
Instead, acceleration-dependent added-mass effects—associated with grain entrainment beneath the foot—dominate transient force responses.
We integrate this force decomposition with a momentum-observer–based estimator that compensates for rigid-body inertia and gravity, and introduce an acceleration-aware weighted regression to account for increased force variance during high-acceleration events.
Together, these methods enable consistent recovery of granular \revise{stiffness parameters}{\resistancename{}} across locomotion conditions, closely matching linear-actuator ground truth. Our results demonstrate that accurate terrain inference during high-speed locomotion requires explicit treatment of acceleration-dependent granular effects, and provide a foundation for robots to characterize complex deformable terrain during dynamic exploration of terrestrial and planetary environments.
\end{abstract}


\section{Introduction}

\begin{figure}[t]
    \centering
    \includegraphics[width=0.90\linewidth]{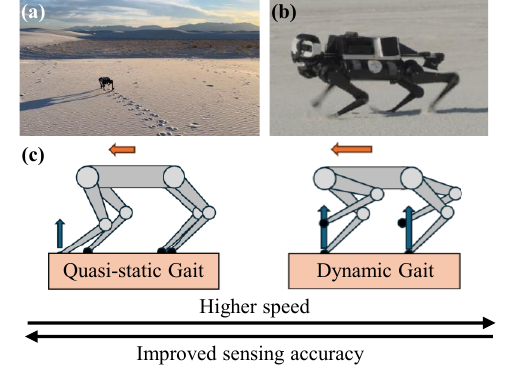}
    \caption{\textbf{Locomotion regimes on deformable terrain and their implications for proprioceptive sensing}. (a) A quadruped robot navigating natural deformable terrain with spatially-varying strength and mechanics.
    (b) Dynamic leg-terrain interaction from a representative stride.
    (c) Locomotion regimes spanning quasi-static gaits (higher sensing fidelity) to dynamic gaits (higher speed), highlighting the trade-off between terrain inference accuracy and locomotion performance. }
    \label{fig:intro}
\end{figure}

Deformable terrain is widespread in natural environments and significantly challenges robot stability and mobility, necessitating focused study of locomotion dynamics and controller design for reliable traversal of yielding substrates~\cite{li2023need}. The mechanical response of natural substrates, such as sand and mud, can vary substantially on compaction state\cite{gravish2014force}, moisture\cite{richefeu2006stress}, and particle size\cite{wang2022micromechanical}.  Robots operating in such environments frequently encounter spatially variable ground conditions \add{(Fig.~\ref{fig:intro}a)}. As a result, locomotion performance depends critically on the robot’s ability to sense and interpret terrain mechanical properties during interaction.
While prior work has advanced adaptive gait control~\cite{hubicki2016tractable, gosyne2018bipedial, liu2025adaptive}, energy-aware control~\cite{lynch2020soft, lynch2024efficient, roberts2021virtual}, and morphological effects~\cite{saranli2001rhex, marvi2014sidewinding, liao2025bio} for deformable substrates, these approaches share a common dependency: controller effectiveness hinges on accurate knowledge of terrain strength and reaction forces at the contact interface. 
Mismatch between assumed and actual terrain properties can degrade performance or lead to failure, underscoring the necessity of terrain awareness—inferring mechanical characteristics of the substrate directly from physical interactions—for safe, reliable, and efficient legged locomotion on deformable environments~\cite{chhaniyara2012terrain, wu2019tactile, chang2017learning, chang2020learning, choi2023learning}.

Recent advances in direct-drive brushless DC (BLDC) actuation provide high torque transparency and low transmission impedance, creating new opportunities for in-situ terrain perception through proprioceptive feedback~\cite{kenneally2018actuator, kenneally2016design}.
Prior work has shown that granular terrain strength can be inferred from internal sensing signals and exploited for locomotion planning~\cite{fulcher2025effect, liu2026scout}, establishing a foundation for terrain characterization without external instrumentation.
This capability has the potential to enable rapid terrain sensing and locomotion adaptation across spatially varying terrain conditions.

However, most existing proprioceptive terrain sensing approaches
 rely on quasi-static granular force models in which resistive forces are assumed to scale primarily with penetration depth
~\cite{albert1999slow,li2013terradynamics, lynch2024efficient}.
Such models neglect inertial and impact effects that arise during high-speed interactions~\cite{goldman2008scaling,qian2013walking,godon2023maneuvering}.  
In actual locomotion, especially under dynamic gaits, robot legs frequently contact the substrate at substantial velocities \add{ (Fig.~\ref{fig:intro}b)}. These interactions introduce significant dynamic effects, where both the magnitude and temporal profile of the terrain reaction forces deviate from quasi-static predictions~\cite{qian2013walking,godon2023maneuvering}, leading to substantial degradation in proprioceptive sensing accuracy during dynamic locomotion~\cite{fulcher2025effect} \add{ (Fig.~\ref{fig:intro}c)}. 
Consequently, simple linear regression of the force–depth curve becomes insufficient when evaluating terrain mechanical properties under dynamic loading conditions.
Impulsive robot–substrate interactions remain largely underexplored in the context of proprioceptive terrain inference. 

To resolve this gap, we investigate dynamics-aware proprioceptive terrain sensing using a direct-drive BLDC-actuated hopper operating on granular media (Sec. \ref{sec:methods}).
We systematically vary impact speed and controller parameters, characterizing terrain reaction forces through both proprioceptive estimation and ground-truth load-cell measurements.
Measurements reveal that under increasing impact velocity, terrain reaction forces deviate substantially from the quasi-static penetration models (Sec. \ref{sec:single_stride}).
\revise{To understand how hopping dynamics affect terrain property inference, we analyze the discrepancy between quasi-static regression and measured force profiles, and isolate the depth, velocity, and acceleration components}{We analyze these discrepancies by decomposing the measured force into depth-, velocity-, and acceleration-dependent components} (Sec. \ref{sec:forward_problem}).
Guided by these observations, we develop a dynamics-aware proprioceptive terrain-sensing framework, that (i) incorporates a momentum observer to account for inertial effects introduced by leg dynamics (Sec. \ref{sec:improved_estimator}), and (ii) employs a granular-physics-informed, acceleration-aware weighted regression to reduce hydrodynamic-like and added-mass contributions, restoring accuracy in terrain parameter inference during high-speed hopping (Sec. \ref{sec:inverse_problem}). 
\delete{Together, these results establish that incorporating impact-aware granular modeling is essential for reliable proprioceptive terrain characterization under dynamic gaits.}
By bridging granular physics with dynamic robot locomotion and proprioceptive sensing, this work advances robot-aided terrain characterization beyond the quasi-static domain, and enables interpretation of complex robot–granular interaction signals during dynamic locomotion. 



\section{Materials and Methods}\label{sec:methods}
Here, we describe the robotic platform, experimental procedures, and data analysis methods to investigate terrain–intruder interactions during hopping. Specifically, Sec. \ref{sec:robot} describes the hopper kinematics and control. Sec. \ref{sec:hopping-exp} describes the experimental setup and data collection procedures. 
Sec. \ref{sec:sate-estimation} describes the state estimation methodology. 

\subsection{Direct-Drive Hopper}\label{sec:robot}
\begin{figure}[t]
    \centering
    \includegraphics[width=0.90\linewidth]{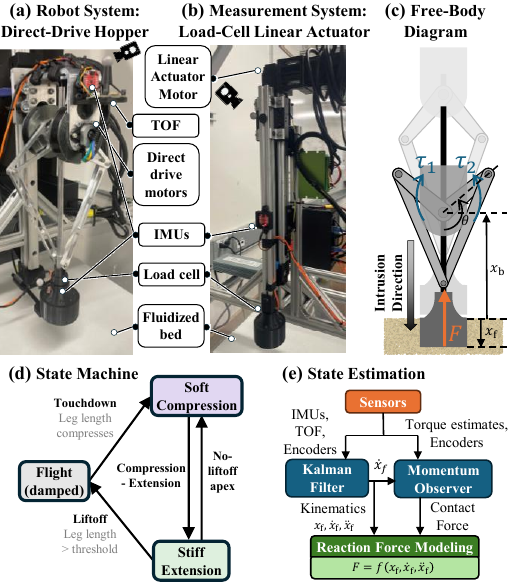}
    \caption{
    (a) Robotic hopper. 
    (b) Load cell instrumented linear actuator for ground truth measurements. 
    (c) Hopper free body diagram. 
    (d) SLIP controller state machine. 
    (e) State estimation pipeline. 
    }
    \label{fig:method}
\end{figure}

A vertically constrained hopper (Fig. \ref{fig:method}a) was used to study proprioceptive granular terrain sensing during dynamic locomotion. The hopper employed a parallel five-bar linkage in which the two upper links were actuated symmetrically to drive the vertical motion of the foot (Fig.~\ref{fig:method}c). A circular foot (80 mm diameter, 50 mm thickness) was attached at the intersection of the two lower linkages via a central bar, enforcing vertical intrusion into the granular terrain. This low-degree-of-freedom (DoF) configuration and simple foot trajectory facilitated systematic variation of impact speed while eliminating effects from body rotation. 

A Raibert-style controller~\cite{raibert1986legged} was used to control the hopper motion. Each hopping cycle was divided into flight and stance phases. During the stance phase, the virtual leg stiffness was modulated from a compliant compression mode to a stiffer extension mode at the end of compression to generate lift-off (Fig.~\ref{fig:method}d). 
This phase-dependent stiffness modulation compensated for energy dissipated during terrain interaction~\cite{lynch2024efficient} by injecting elastic potential energy through the virtual spring. During flight, the leg reset to its compression neutral length for the next stance; additional damping was applied to suppress oscillations in the absence of terrain loading.

The joints were actuated using direct-drive (\textit{i.e.,} gearless) motors (CubeMars, R80 KV110), which provided high force transparency~\cite{kenneally2018actuator} and enabled estimation of ground reaction forces from actuator current measurements. Joint position and current were logged from motor encoders at a sampling frequency of 1kHz. Joint torque was estimated from the joint current, enabling the computation of end-effector forces using Jacobian:
\begin{equation}
\bm{F} = \bm{J}^{-\top}\bm{\tau}
\label{eqn:quasi_static}
\end{equation}
where $\bm{J}$ is the leg Jacobian and $\bm{\tau} = \mathbf{K}_\mathrm{t}\bm{I}$ denotes the joint torque vector estimated from measured joint currents $\bm{I}$ using the torque-constant $\mathbf{K}_\mathrm{t}$.

To evaluate the accuracy of the proprioceptive force estimates, a single-axis compression load cell (50 kg rated capacity) was mounted on the foot and interfaced with a signal amplifier (ATO-LCTR-DY510) and a 16-bit ADC (ADS1115).  
In addition, two inertial measurement units (IMUs; ISM330DHCX) were mounted on the body and the foot, respectively, to measure linear acceleration.
A downward-facing time-of-flight (ToF) sensor (AFBR-S50) was installed on the body to measure body height. 

\subsection{Hopping Experiments for Proprioceptive Terrain Characterization}
\label{sec:hopping-exp}
Experiments were conducted on a granular bed composed of \SI{300}{\micro\meter} glass beads, which qualitatively resembles natural sand, while its uniform size and shape promote consistent mechanical behavior~\cite{hill2005scaling}.
Prior to each trial, the bed was reset using a fluidized-bed preparation technique~\cite{jin2019preparation}, producing a level surface and repeatable volume fraction. \add{This preparation ensured repeatable granular resistance, $k$, the penetration resistance per unit depth widely used to characterize deformable terrain strength~\cite{stone2004local,qian2015principles}}.

To ensure interaction with undisturbed terrain, each hopping trial was performed on a freshly prepared surface.
The robot was released from a prescribed height, and only the first complete hopping stride—from pre-touchdown flight to post-liftoff—was analyzed to avoid terrain compaction from prior contacts.
Five impact conditions were tested\revise{, spanning measured touchdown speeds $v_{\mathrm{TD}}\approx 0.2$--$1.2\,\mathrm{m/s}$}{ at nominal touchdown speeds 
$v_{\mathrm{TD}} = 0.22$, $0.51$, $0.83$, $1.01$, and $1.22~\mathrm{m/s}$} \delete{(the lowest-speed condition used an $\sim$2\,mm release)}.
The extension stiffness was fixed at 5.00~N/cm, while compression stiffness varied among 2.50, 3.75, and 5.00~N/cm. Neutral lengths were held constant. Five trials per configuration were performed, yielding 60 total trials. 
All onboard sensors—including motor torque estimates, encoders, IMUs, the ToF sensor, and the load cell—recorded synchronized measurements at 1 kHz during each trial to capture the transient robot–terrain interaction dynamics. 
Ground-truth body and foot positions were tracked using the motion-capture system (OptiTrack Prime 13W, 240 FPS). In addition, Hopper dynamics and foot-ground interactions were recorded from sideview (GoPro HERO 12, 240 FPS).


To obtain ground truth terrain mechanics, we used a linear actuator (Heechoo) mounted above the fluidized bed to perform penetration experiments (Fig.~\ref{fig:method}b). During each trial, the linear actuator drove the same hopper foot vertically into the granular bed, while a load cell measured the resulting resistive forces under high-speed intrusion. 
Fifty intrusion speeds \add{ranging from 0.022 to 1.1~m/s in 0.022~m/s increment} were tested, with a fixed 5~cm maximum depth. Three trials per speed yielded 150 total trials.
Motion capture provided independent measurements of intrusion depth, while an IMU measured acceleration and, via sensor fusion, estimated velocity.

\subsection{State Estimation}\label{sec:sate-estimation}
To obtain inertia-compensated proprioceptive force estimates, we used a discrete-time Kalman filter to estimate system states from onboard sensing and a momentum observer to account for inertial effects in robot–terrain interaction (Fig.~\ref{fig:method}e).
\paragraph{Kalman Filter}
State estimation is performed using a discrete-time Kalman filter~\cite{kalman1960new}. The system kinematics are modeled in discrete time as,
\begin{equation}
\bm{x}_{k+1} = \bm{A}_k\bm{x}_k + \bm{B}_k\bm{u}_k + \bm{w}_k,
\end{equation}
where the state vector is defined as
$\bm{x}_k \coloneqq
[\, x_{\mathrm{b},k},\ \dot{x}_{\mathrm{b},k},\ x_{\mathrm{f},k},\ \dot{x}_{\mathrm{f},k} \,]^{\top}$, with $x_{\mathrm{b},k}$ and $x_{\mathrm{f},k}$ denoting the vertical positions of the body and the foot, respectively, and $\dot{x}_\mathrm{b}$ and $\dot{x}_\mathrm{f}$ their corresponding velocities (Fig.~\ref{fig:method}c).
The input vector is
$\bm{u}_k \coloneqq
[\, a_{\mathrm{b},k},\ a_{\mathrm{f},k} \,]^{\top}$,
where $a_\mathrm{b}$ and $a_\mathrm{f}$ the linear accelerations measured by the IMU.
The state transition matrix $\bm{A}_k$ and input matrix $\bm{B}_k$ are given by
\begin{equation}
\bm{A}_k =
\begin{bmatrix}
1 & \Delta t_k & 0 & 0 \\
0 & 1 & 0 & 0 \\
0 & 0 & 1 & \Delta t_k \\
0 & 0 & 0 & 1
\end{bmatrix},\quad
\bm{B}_k =
\begin{bmatrix}
\frac{1}{2}\Delta t_k^2 & 0 \\
\Delta t_k & 0 \\
0 & \frac{1}{2}\Delta t_k^2 \\
0 & \Delta t_k
\end{bmatrix},
\end{equation}
where $\Delta t_k$ denotes the discrete-time sampling interval.
And $\mathbf{w}_k$ represents zero-mean process noise induced by IMU measurement uncertainty.

The measurement model is given by
\begin{equation}
\bm{z}_k = \bm{H}_k \bm{x}_k + \bm{v}_k ,
\end{equation}
where
\begin{equation}
\bm{z}_k =
\begin{bmatrix}
z_{k,1} \\
z_{k,2} \\
z_{k,3}
\end{bmatrix}
=
\begin{bmatrix}
x_{\mathrm{b},k} \\
x_{\mathrm{b},k}-x_{\mathrm{f},k} \\
\dot{x}_{\mathrm{b},k}-\dot{x}_{\mathrm{f},k}
\end{bmatrix}
+\bm{v}_k ,
\end{equation}
\begin{equation}
\bm{H}_k =
\begin{bmatrix}
1 & 0 & 0 & 0 \\
1 & 0 & -1 & 0 \\
0 & 1 & 0 & -1
\end{bmatrix},
\end{equation}
with $z_{k,1}$ measured by the ToF sensor; $z_{k,2}$ and $z_{k,3}$ corresponding to the body–foot relative displacement and its rate obtained from the encoder; and $\mathbf{v}_k$ denoting zero-mean measurement noise.


\paragraph{Momentum Observer}
\revise{The estimated kinematics are used to reconstruct the foot–terrain contact force via a momentum observer.}
{To capture dynamics omitted in the quasi-static reconstruction in~\eqref{eqn:quasi_static}, we formulate a momentum observer~\cite{haddadin2017robot, morlando2021whole}.}
\revise{Using generalized coordinates $\bm{q}=[x_f,\ \theta]^\top$, where $\theta$ is the joint angle shared by the two motors under symmetric actuation,}{Under symmetric actuation, the two motors are represented by two scalars: the joint coordinate $\theta$ and input torque $\tau$. Using 
the generalized coordinates $\bm{q}=[x_f,\ \theta]^\top$, }
the dynamics are written as
\begin{equation}
\bm{M}(\bm{q})\,\ddot{\bm{q}}
+
\bm{C}(\bm{q},\dot{\bm{q}})\,\dot{\bm{q}}
+
\bm{g}(\bm{q})
=
\mathbf{S}^\top\,\tau
+
\mathbf{J}_{\mathrm{c}}^{\top}\,F_{\mathrm{c}},
\label{eq:robot_dynamics}
\end{equation}
where \delete{$\tau$ is the scalar input torque and }$F_{\mathrm{c}}$ is the scalar vertical foot--terrain contact force. 
With this coordinate choice, the selection matrix and contact Jacobian reduce to constant row vectors, $\mathbf{S}=\begin{bmatrix}0 & 1\end{bmatrix}$ and $\mathbf{J}_{\mathrm{c}}=\begin{bmatrix}1 & 0\end{bmatrix}$, and are therefore independent of $\mathbf{q}$. 
Here, $\mathbf{J}_{\mathrm{c}}(\mathbf{q})$ denotes the \emph{contact Jacobian}, which is distinguished from the Jacobian used for quasi-static force reconstruction in~\eqref{eqn:quasi_static}.

Using this formulation, we solve for the foot vertical acceleration $\ddot{x}_f$ from~\eqref{eq:robot_dynamics}, yielding
\begin{equation}
M_\mathrm{f}(\theta)\,\ddot{x}_\mathrm{f} + M_\mathrm{f}(\theta)\,g = F_\mathrm{c} - \beta(\theta)\,\tau - C(\theta)\,\dot{\theta}^2,
\end{equation}
This expression relates the vertical foot dynamics to the foot--terrain contact force and the applied joint torque. 
When the linkage inertia and Coriolis/centrifugal terms are neglected, the model reduces to the quasi-static proprioceptive formulation in~\eqref{eqn:quasi_static}.
As acceleration measurements are noisy and not directly used in the momentum observer, we reformulate the estimator in terms of the foot momentum
\begin{equation}
p_{\mathrm{f}} \coloneqq M_{\mathrm{f}}(\theta)\,\dot{x}_{\mathrm{f}} .
\end{equation}
Taking the time derivative and substituting the dynamics yields the contact force in the form
\begin{equation}
F_{\mathrm{c}} = \dot{p}_{\mathrm{f}} - \psi(\theta,\dot{\theta},\dot{x}_{\mathrm{f}},\tau),
\label{eq:Fc_momentum_form}
\end{equation}
where
\begin{equation}
\psi(\cdot) \coloneqq 
\frac{\partial M_{\mathrm{f}}(\theta)}{\partial \theta}\,\dot{\theta}\,\dot{x}_{\mathrm{f}}
- M_{\mathrm{f}}(\theta)\,g
- \beta(\theta)\,\tau
- C(\theta)\,\dot{\theta}^2 .
\end{equation}

Following~\cite{haddadin2017robot}, the contact force was estimated using a first-order residual filter
\begin{equation}
\dot{r} = k\big(F_{\mathrm{c}}-r\big),
\label{eq:residual_filter}
\end{equation}
which can be implemented equivalently via an internal momentum estimate $\hat{p}_{\mathrm{f}}$:
\begin{equation}
r = k\big(p_{\mathrm{f}}-\hat{p}_{\mathrm{f}}\big),
\qquad
\dot{\hat{p}}_{\mathrm{f}} = \psi(\cdot) + r .
\label{eq:momentum_observer}
\end{equation}
Encoder measurements provided $\theta$ and $\dot{\theta}$, and $\dot{x}_{\mathrm{f}}$ was obtained from the Kalman filter. The contact force was then recovered from the residual $r$.


\section{Results}
This section reports force measurements during dynamic hopping, isolates factors influencing ground reaction forces, and develops a granular-physics framework to infer terrain properties from proprioceptive data.
Sec. \ref{sec:single_stride} presents a representative hopping stride, illustrating kinematic states and terrain contact forces. Sec. \ref{sec:forward_problem} explores hydrodynamic-like effects under high-speed intrusion and their manifestation in the contact force.
Sec. \ref{sec:improved_estimator} introduces a momentum-observer estimator to compensate for rigid body inertia, recovering terrain forces and improving proprioceptive estimates.
Finally, Sec.~\ref{sec:inverse_problem} integrates the robot–granular interaction model with improved force estimates to infer granular terrain properties during high-speed hopping.

\subsection{Hopper Dynamics and Terrain Reaction Force During a Single Cycle}
\label{sec:single_stride}
\begin{figure}[htbh!]
    \centering
    \includegraphics[width=0.90\linewidth]{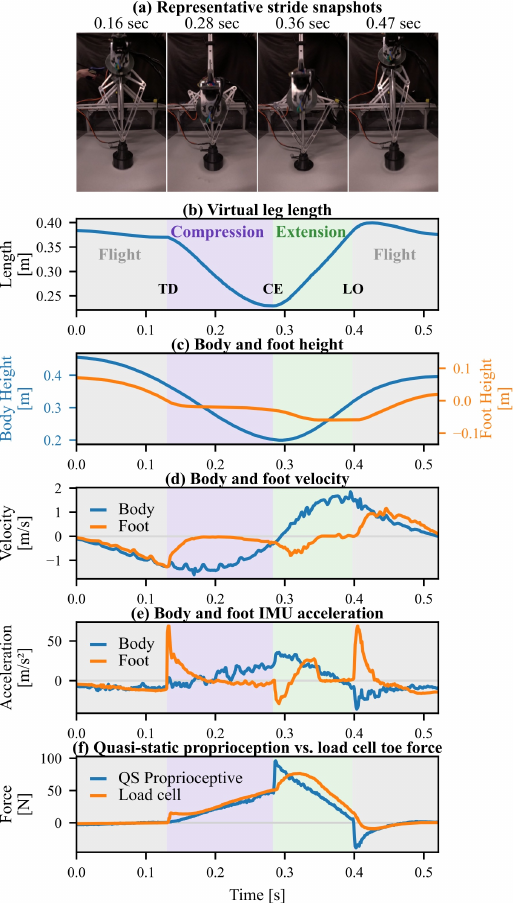}
    \caption{
    \textbf{Representative dynamic hopping stride on granular medium.}
    (a) Snapshots of the hopper during a representative stride: initial compression after touchdown, maximum compression, deeper penetration induced by the stiffer extension phase, and post-liftoff.
    (b) Virtual leg length, defined as the body–foot height difference, measured by motor encoder. TD: touchdown; CE: compression–extension transition; LO: liftoff.
    (c, d) Body and foot height and velocity, respectively, measured via motion capture (MoCap).
    (e) Body and foot acceleration measured by onboard IMUs.
    (f) Quasi-static proprioceptive estimate~\eqref{eqn:quasi_static} compared with load-cell ground-truth measurements.
    }
    \label{fig:single_stride}
\end{figure}

We examined the  evolution of the hopper dynamics and ground reaction forces during a representative stride cycle on undisturbed sand (Fig.~\ref{fig:single_stride}).
The cycle was segmented into pre-touchdown flight, compression, extension, and post-liftoff flight.

At touchdown (Fig.~\ref{fig:single_stride}, $t = 0.13$ s), the virtual leg length began decreasing as the compression phase initiated (Fig.~\ref{fig:single_stride}b).
Upon contact, the body continued descending while the foot height briefly plateaued before further penetration (Fig.~\ref{fig:single_stride}c).
Unlike rigid-ground hopping, where the foot velocity drops instantaneously to zero upon touchdown, the foot velocity decelerated over approximately 50 ms before reaching rest (Fig.~\ref{fig:single_stride}d) due to granular medium yielding.
The foot IMU exhibited a sharp upward acceleration transient at impact (Fig.~\ref{fig:single_stride}e), corresponding to rapid momentum exchange with the substrate.
The load-cell force increased immediately after contact and exhibited an early peak (Fig.~\ref{fig:single_stride}f, orange curve).
This peak was lower and broader than the sharp, high-magnitude impact forces typical of rigid-ground contact, reflecting pronounced energy dissipation in the deformable terrain.

During compression (Fig.~\ref{fig:single_stride}, $t = 0.13 - 0.28$ s), the virtual leg shortened until reaching its minimum (Fig.~\ref{fig:single_stride}b).
The foot penetrated deeper into the sand while the body descended toward its minimum height (Fig.~\ref{fig:single_stride}c).
Both body and foot velocities approached zero near maximum leg compression (Fig.~\ref{fig:single_stride}d), while body acceleration increased and remained elevated due to the rising controller-generated leg force (Fig.~\ref{fig:single_stride}e).
The contact force increased throughout compression (Fig.~\ref{fig:single_stride}f); unlike rigid-ground hopping, where a large ground-reaction force rapidly halts the foot and balances the applied leg force, the substrate yielded nonlinearly under load.

At the compression–extension transition (Fig.~\ref{fig:single_stride}, $t=0.28$~s), the controller increased the virtual leg stiffness, initiating leg extension (Fig.~\ref{fig:single_stride}b).
As extension began, the body height increased while the foot continued descending due to substrate deformation, creating a transient divergence (Fig.~\ref{fig:single_stride}c).
The body velocity rapidly became positive, whereas the foot velocity remained negative before approaching zero before reversing (Fig.~\ref{fig:single_stride}d).
A second pronounced acceleration transient was observed at the stiffness switch (Fig.~\ref{fig:single_stride}e).
The contact force increased after the stiffness transition, reaching its maximum during extension (Fig.~\ref{fig:single_stride}f), coinciding with maximum penetration depth.
This behavior differed from rigid-ground hopping, where force would decrease once upward acceleration began.
The deformable substrate continued yielding under increased load, causing further penetration and energy dissipation.

Upon liftoff (Fig.~\ref{fig:single_stride}, $t= 0.47$ s), the controller switched to flight, and the leg returned toward its neutral length in preparation for the next stride (Fig.~\ref{fig:single_stride}b).
The foot lifted off below the original sand surface, indicating residual substrate deformation (Fig.~\ref{fig:single_stride}c).
A third acceleration transient occurred at liftoff (Fig.~\ref{fig:single_stride}e), reflecting rapid unloading and leg retraction.

\subsection{Decomposing Terrain Reaction Force in Dynamic Intrusion}\label{sec:forward_problem}


\begin{figure}[t]
    \centering
    \includegraphics[width=0.90\linewidth]{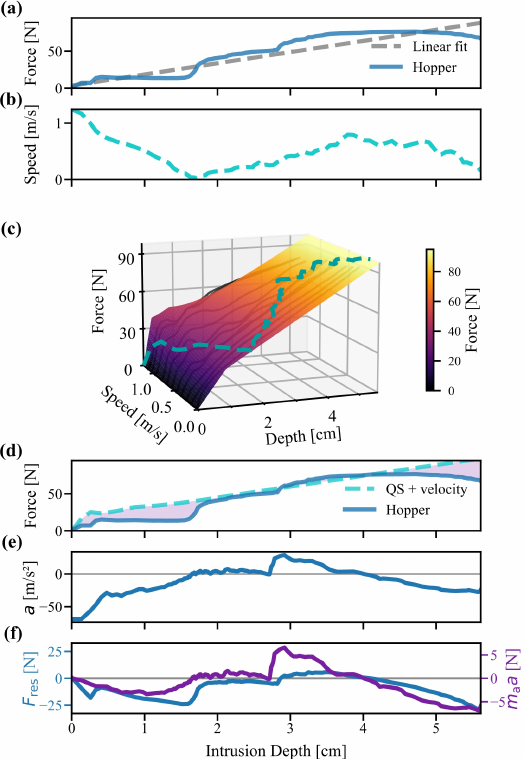}
    \caption{\textbf{Terrain force decomposition under high-speed intrusion.}
    (a) Force–depth with linear fit.
    (b) Intrusion speed–depth.
    (c) Depth–speed force map from constant-speed linear-actuator intrusions; gray: representative trials, cyan: hopper trajectory projected for prediction.
    (d) Measured force vs map prediction.
    (e) Foot acceleration. Positive represents downward direction. 
    (f) $F_{\mathrm{res}}$ and added-mass term $m_\mathrm{a}a$.
    }
    \label{fig:forward_problem}
\end{figure}

In this section, we integrate granular physics models to develop a terrain-force prediction framework using ground reaction forces measured by the load cell during hopping.

\subsubsection{Quasi-static reference}
Granular intrusion at low speed is commonly modeled under quasi-static penetration as a depth-dependent resistive force that scales approximately linearly with penetration depth beyond a shallow transient regime~\cite{kang2018archimedes}.
Most robotic terrain-sensing work adopts this quasi-static (QS) formulation to interpret terrain properties~\cite{qian2019rapid, ruck2024downslope}.
We therefore first compared the hopper’s depth–force measurements to a linear depth-only fit; the systematic deviation indicates that depth alone cannot capture the force in the dynamic regime (Fig.~\ref{fig:forward_problem}(a)).


\subsubsection{Velocity-dependent effects}
We hypothesized that the mismatch arose from velocity-dependent inertial effects during high-speed intrusion.
Prior studies~\cite{goldman2008scaling, qian2013walking} showed that granular resistive forces scale quadratically with intrusion speed once impact exceeds a material-dependent inertial threshold, which for our medium is $\sqrt{2d_\mathrm{g}g}\approx0.08$~m/s~\cite{roth2021intrusion}.
Because touchdown speeds in our hopping experiments exceeded this threshold (Fig.~\ref{fig:forward_problem}b), the interaction lay in the inertial regime.

To characterize velocity dependence, we performed constant-speed intrusions from 0.022 to 1.1 m/s, with an increment of 0.02 m/s (Fig.~\ref{fig:forward_problem}c, transparent curves). 
Using these data, we fit a compact parametric terrain model via ordinary least squares and constructed a continuous depth–speed force surface (Fig.~\ref{fig:forward_problem}c, surface color).


Incorporating velocity dependence improved agreement relative to the QS model, but substantial residuals remained (Fig.~\ref{fig:forward_problem}d), particularly (i) shortly after touchdown and (ii) post compression–extension transition.
Both intervals coincided with rapid foot deceleration (Fig.~\ref{fig:forward_problem}e), suggesting an additional acceleration-dependent contribution.

\subsubsection{Acceleration-dependent added mass}
Granular impact studies report that high-speed intruders entrain a conical region of grains beneath the foot, which behaves as an effective added mass~\cite{aguilar2016robophysical}. This mechanism introduces acceleration-dependent inertial forces beyond depth- and velocity-dependent resistance.
Building on this framework, We modeled the terrain reaction force as:
\begin{equation}
F_\mathrm{g}(z,\dot{z},\ddot{z})
= F_\mathrm{d}(z) + \frac{\mathrm{d}}{\mathrm{d}t}[m_\mathrm{a}(z)\dot{z}]=F_\mathrm{d}(z)+\frac{\partial{m_\mathrm{a}}}{\partial{z}}\dot{z}^2+m_\mathrm{a}\ddot{z},
\label{eq:terrain_force}
\end{equation}
where $z=\max(0, -x_\mathrm{f})$ denotes the penetration depth (positive downward), $m_\mathrm{a}$ denotes the added mass.

This formulation separates inertial effects into: (1) a velocity-dependent quadratic drag term proportional to $\dot{z}^2$, and (2) an acceleration-dependent added-mass term $\ddot{z}$ arising from accelerating the granular cone.
Under constant-speed intrusion, $\ddot{z}=0$, and only the $\dot{z}^2$ contribution remains.
However, during hopping, rapid deceleration and acceleration occurred at touchdown and at the compression–extension transition, activating the $m_a\ddot{z}$ term.
The added mass grew during the near-surface transient regime and became approximately constant once the steady regime is reached.


We estimated the added-mass profile by integrating the fitted added-mass gradient with respect to depth, and computed an acceleration-induced contribution using the estimated added mass and the measured acceleration (Fig.~\ref{fig:forward_problem}f, purple). 
The resulting added-mass term reproduced the dominant timing and shape of the residual force during both transient intervals (Fig.~\ref{fig:forward_problem}f, blue).
Although magnitudes differed modestly, likely due to uncertainty in gradient integration and bulk-model fitting, the agreement supports acceleration-induced added mass as a primary contributor to dynamic terrain reaction forces during hopping.

This decomposition established a physics-based representation of terrain reaction forces under dynamic intrusion.
In Sec.~\ref{sec:inverse_problem}, we leverage this representation to infer terrain properties from proprioceptive force measurements.

\subsection{Proprioceptive Terrain Reaction Force Estimation via Momentum Observer}
\label{sec:improved_estimator}
To infer terrain properties from onboard sensing alone, proprioceptive force estimates must accurately capture transient interaction forces.
However, the quasi-static (QS) estimator~\eqref{eqn:quasi_static} exhibited substantial mismatch with load-cell ground truth, particularly during rapid events (Fig.~\ref{fig:single_stride}f). It failed to reproduce (i) the initial force surge at touchdown and (ii) the two force peaks associated with controller mode transitions.

These discrepancies aligned with three pronounced foot-IMU acceleration transients (Fig.~\ref{fig:single_stride}(e))—impact, the compression–extension transition, and liftoff—indicating rapid momentum exchange neglected by the quasi-static model.
Conceptually, the quasi-static estimator treats the body–foot spring interaction as the terrain force; however, when the foot accelerates, foot inertia causes a mismatch between the body–foot interaction and the true terrain contact force, biasing the inferred terrain force—especially during impact and phase transitions.
This motivates accurate state (and acceleration) estimation to improve proprioceptive terrain inference.

\begin{figure}[t]
    \centering
    \includegraphics[width=0.90\linewidth]{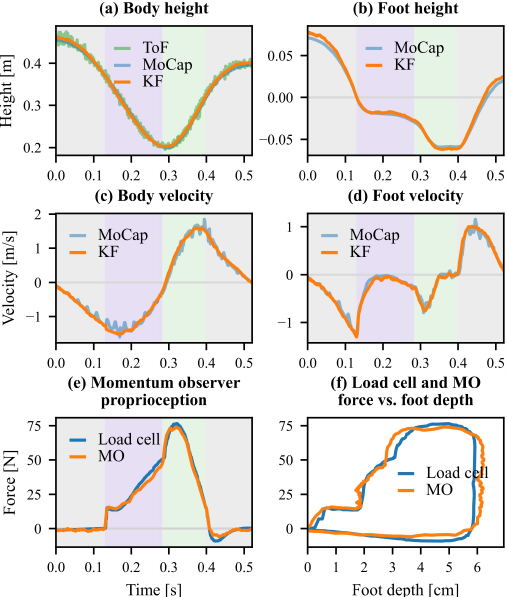}
    \caption{\textbf{Kinematic state estimation and MO-based force inference.}
    (a-d) Body and foot height (a,b) and velocity (c,d) estimated via the onboard-sensing-based Kalman filter, compared with MoCap measurements.
    (e) Toe force estimated by the momentum observer using Kalman-filter-estimated kinematics.
    (f) MO force vs. KF-estimated depth, compared with load-cell force vs. MoCap depth.}
    \label{fig:improved_estimator}
\end{figure}

To address this limitation, we implemented a momentum-observer framework driven by onboard state estimation (Fig.~\ref{fig:improved_estimator}), with its kinematic inputs provided by a Kalman filter that fuses ToF, encoder, and IMU measurements to estimate body and foot motion.
The ToF sensor measured body height over granular medium but was noisy and unsuitable for direct height or velocity use.
The encoder provided high‑fidelity body–foot displacement with low assumed covariance, and the IMU, despite noise, tracked body and foot kinematics to constrain the filter between updates.
This fusion yielded reliable onboard kinematic estimates, validated against motion capture(Fig.~\ref{fig:improved_estimator}a-d).
The body height closely tracked the MoCap data, effectively denoising the ToF measurements, while the foot height also aligned well with MoCap.
More importantly, the velocity, obtained through Kalman filter, matched the MoCap velocity with high accuracy.

Using the estimated foot velocity, the momentum observer compensated for inertia and gravity to recover the external terrain reaction force.
The observer-based estimate closely matched the load-cell force throughout the stride, including impact and phase transitions, substantially improving upon the QS estimate (Fig.~\ref{fig:single_stride}e).
Finally, combining Kalman-filtered foot height with the force estimates, we expressed force as a function of intrusion depth (Fig.~\ref{fig:improved_estimator}f) and found that, despite residual errors, the primary terrain mechanics were captured relative to load-cell and MoCap ground truth, establishing momentum-observer–based proprioception as a viable foundation for terrain characterization during dynamic hopping.

\subsection{Dynamic Terrain Reaction Force Characterization for Robot–Terrain Interaction}
\label{sec:inverse_problem}
In this section, we investigate whether \emph{proprioceptively estimated contact forces} can be used to recover an empirical terrain reaction model $F(z,\dot{z},\ddot{z})$, and evaluate the performance of three treatments:
\begin{itemize}
    \item \textbf{w/o MO, w/o GD}: quasi-static (QS) proprioception only ~\eqref{eqn:quasi_static} with linear fitting to estimate granular \resistancename{} without considering granular dynamics;
    \item \textbf{w/ MO, w/o GD}: momentum observer (MO) corrected proprioception, with linear fitting.
    \item \textbf{w/ MO, w/ GD}: momentum observer corrected proprioception with acceleration-aware weighted least squares (WLS) fitting that accounts for acceleration-induced granular dynamics (GD).
\end{itemize}

\subsubsection{Momentum observer improves force estimation and resulting granular strength estimation}
\begin{figure}[t]
    \centering
    \includegraphics[width=0.90\linewidth]{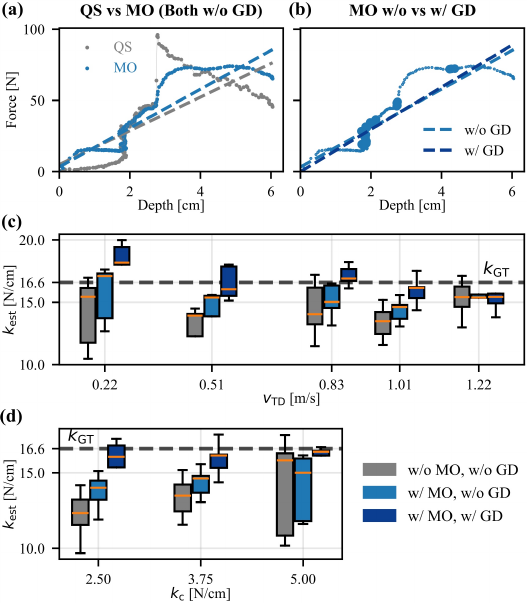}
    \caption{\textbf{\revise{Terrain stiffness}{Granular \resistancename{}} estimation across three treatments.}
    (a) Representative depth--force curve comparing w/o MO, w/o GD (QS; OLS fit) versus w/ MO, w/o GD (MO-corrected; OLS fit).
    (b) Representative MO-based fit comparing w/ MO, w/o GD (OLS) versus w/ MO, w/ GD (acceleration-aware WLS; points weighted in the regression).
    (c,d) Estimated \revise{stiffness}{\resistancename{}} for w/o MO, w/o GD (gray), w/ MO, w/o GD (light blue), and w/ MO, w/ GD (dark blue), shown versus (c) touchdown speed and (d) compression stiffness.
    In (c), $k_c=3.75\,\mathrm{N/cm}$; in (d), $v_{\mathrm{TD}}\approx1.0\,\mathrm{m/s}$.
    Points are means over $n=5$ trials; error bars indicate $\pm$SEM.
    Dashed lines in (c,d) denote $k_{\mathrm{GT}}$, the linear-actuator ground-truth granular \delete{terrain} \revise{stiffness}{\resistancename{}}.}
    \label{fig:inverse_problem}
\end{figure}

We first compared terrain characterization using momentum observer based force estimate against the quasi-static estimator~\eqref{eqn:quasi_static}.
Across touchdown speeds (fixed $k_c$) and across controller stiffnesses (fixed $v_{\mathrm{TD}}$), the MO-corrected estimates (Fig.~\ref{fig:inverse_problem}c,d, \textbf{w/ MO, w/o GD}) systematically lay closer to the linear-actuator ground truth $k_{\mathrm{GT}}$ than the QS estimator (Fig.~\ref{fig:inverse_problem}c,d, \textbf{w/o MO, w/o GD})\add{, reducing trial-level RMSE from $4.84$ to $3.47~\mathrm{N/cm}$}.
In contrast, the uncorrected QS force–depth relationship was strongly nonlinear with substantial within-group variability, undermining linear \revise{stiffness}{\resistancename{}} regression.
This improvement indicated that compensating for rigid-body inertia and gravity reduced systematic bias in proprioceptive \revise{stiffness}{\resistancename{}} estimation.
Nevertheless, residual deviations from $k_{\mathrm{GT}}$ remained, including at high touchdown speeds with large accelerations.

\subsubsection{Acceleration-aware weighting enables identification of granular terrain mechanics during dynamic hopping}
We next examine how explicitly accounting for granular dynamics influences the inferred terrain parameters.
As shown in Sec. \ref{sec:forward_problem}, residual force discrepancies 
correlated with large foot accelerations. To account for this, we applied an 
acceleration-dependent weighting in regression that down-weights 
samples collected during large $|\ddot z|$. Specifically, we assign each sample $i$ an inverse-variance weight
\begin{equation}
    w_i = \frac{1}{\sigma_i^2}, 
    \qquad 
    \sigma_i = \sigma_{\mathrm{good}} 
    + (\sigma_{\mathrm{bad}} - \sigma_{\mathrm{good}})\, s_i,
\end{equation}
where
\begin{equation}
    s_i = \left(1 + 
    \exp\!\left[-k\left(|\ddot z_i| - a_0\right)\right]\right)^{-1}.
\end{equation}
The sigmoid mapping $s_i$ increases the assigned variance once $|\ddot z|$ exceeds a threshold $a_0$, \revise{reducing the influence of high-acceleration samples}{treating high-acceleration samples as
higher-variance measurements} while retaining them in the regression.

\revise{Across touchdown speeds at fixed $k_c$, the acceleration-aware estimation using granular dynamics  (Fig.~\ref{fig:inverse_problem}c, \textbf{w/ MO, w/ GD}) yielded $k_{\mathrm{est}}$ values that tracked the linear-actuator reference $k_{\mathrm{GT}}$ closely.}{Across the touchdown-speed sweep at fixed $k_c$ and the stiffness sweep at fixed $v_{\mathrm{TD}}$ (Fig.~\ref{fig:inverse_problem}c, d), the acceleration-aware estimate ( \textbf{w/ MO, w/ GD}) tracked $k_{\mathrm{GT}}$ more closely and further reduced trial-level RMSE from $3.47~\mathrm{N/cm}$ to $2.35~\mathrm{N/cm}$.}
A similar improvement was observed when varying $k_c$ at fixed $v_{\mathrm{TD}}$ (Fig.~\ref{fig:inverse_problem}d). 
These results showed that down-weighting high-acceleration samples improves granular terrain parameter identification under dynamic hopping conditions.

In summary, compensating for inertia and gravity using a momentum observer is necessary to obtain consistent proprioceptive \revise{stiffness}{\resistancename{}} estimates across dynamic locomotion regimes.
Accounting for acceleration-dependent error in regression further improves agreement with ground truth.
Together, MO-based force estimation combined with acceleration-aware fitting enables robust onboard characterization of granular \revise{stiffness}{\resistancename{}} during high-speed hopping.

\section{Conclusion}
In this work, we presented a dynamics-aware proprioceptive framework for terrain characterization on dry granular media using onboard sensing alone.
Controlled high-speed hopping experiments showed that quasi-static terrain-\revise{stiffness}{\resistancename{}} estimation broke down during impact and controller-induced stiffness transitions, where force discrepancies grow with foot acceleration.
By incorporating granular physics models, we identified velocity- and acceleration-dependent effects as dominant contributors to transient force responses.
A Kalman-filter state estimator and momentum observer then recovered contact forces accurately from onboard sensing, enabling terrain property identification without external instrumentation.
Furthermore, incorporating acceleration-aware weighted regression improved \revise{stiffness}{\resistancename{}} identification across touchdown speeds and leg compliances, yielding consistent agreement with linear-actuator ground truth.
Together, these results show that explicit treatment of inertial granular effects is necessary for reliable terrain inference during dynamic locomotion.

\revise{Looking forward, this framework opens several avenues for extending dynamics-aware proprioception beyond the present setting.
First, the proposed approach can be generalized to diverse robotic leg morphologies, multi-degree-of-freedom contacts, and substrates with cohesion or fluid saturation, where geometry- and direction-dependent force representations are required.
Such extensions could support real-time terrain characterization and adaptive control for legged robots operating in complex terrestrial and planetary environments.
In addition, leveraging dense step-wise interaction data generated during locomotion may enable learning-based or hybrid physics–data models that capture context-dependent contact phenomena while preserving physical interpretability.
By treating each locomotion step as a sensing experiment, robots could collect spatially resolved terrain data during exploration and continuously update internal terrain representations, constructing and refining world models through interaction.}
{Looking forward, this framework can be extended to diverse leg morphologies, multi-degree-of-freedom contacts, and cohesive or fluid-saturated substrates, where geometry- and direction-dependent force representations may be required.
Such extensions could support real-time terrain characterization, adaptive control, and hybrid physics--data models for complex terrestrial environments.
By treating each step as a sensing experiment, robots could build and refine spatially resolved terrain representations through interaction.}

\IEEEpeerreviewmaketitle


%


\ifreview\else
\section*{Acknowledgment}
\revise{The authors thank Jordan Westphal for simulations and discussions. This work is supported by the National Science Foundation (NSF) CAREER Award \#2240075, the NASA Planetary Science and Technology Through Analog Research (PSTAR) program, Award \# 80NSSC22K1313, the NASA Lunar Surface Technology Research (LuSTR) program, Award \# 80NSSC24K0127, National Science Foundation (NSF) Foundational Research in Robotics (FRR) program, Award \#2529696, and the NASA Mars Exploration Program (MEP) Technology Development Funding.}
{The authors thank Jordan Westphal for assistance with simulation. This work was supported by NSF CAREER \#2240075, NSF FRR \#2529696, NASA PSTAR \#80NSSC22K1313, NASA LuSTR \#80NSSC24K0127, and NASA MEP Technology Development Funding.}
\fi
\ifCLASSOPTIONcaptionsoff
  \newpage
\fi



\bibliographystyle{IEEEtran}
\bibliography{references}

@article{stone2004local,
  title={Local jamming via penetration of a granular medium},
  author={Stone, MB and Barry, R and Bernstein, DP and Pelc, MD and Tsui, YK and Schiffer, P},
  journal={Physical Review E—Statistical, Nonlinear, and Soft Matter Physics},
  volume={70},
  number={4},
  pages={041301},
  year={2004},
  publisher={APS}
}

@article{qian2015principles,
  title={Principles of appendage design in robots and animals determining terradynamic performance on flowable ground},
  author={Qian, Feifei and Zhang, Tingnan and Korff, Wyatt and Umbanhowar, Paul B and Full, Robert J and Goldman, Daniel I},
  journal={Bioinspiration \& biomimetics},
  volume={10},
  number={5},
  pages={056014},
  year={2015},
  publisher={IOP Publishing}
}

@article{chhaniyara2012terrain,
  title={Terrain trafficability analysis and soil mechanical property identification for planetary rovers: A survey},
  author={Chhaniyara, S and Brunskill, C and Yeomans, B and Matthews, MC and Saaj, C and Ransom, S and Richter, L},
  journal={Journal of Terramechanics},
  volume={49},
  number={2},
  pages={115--128},
  year={2012},
  publisher={Elsevier}
}

@article{saranli2001rhex,
  title={RHex: A simple and highly mobile hexapod robot},
  author={Saranli, Uluc and Buehler, Martin and Koditschek, Daniel E},
  journal={The International Journal of Robotics Research},
  volume={20},
  number={7},
  pages={616--631},
  year={2001},
  publisher={SAGE Publications}
}

@article{albert1999slow,
  title={Slow drag in a granular medium},
  author={Albert, R and Pfeifer, MA and Barab{\'a}si, A-L and Schiffer, P},
  journal={Physical review letters},
  volume={82},
  number={1},
  pages={205},
  year={1999},
  publisher={APS}
}

@inproceedings{gosyne2018bipedial,
  title={Bipedial locomotion up sandy slopes: Systematic experiments using zero moment point methods},
  author={Gosyne, Jonathan R and Hubicki, Christian M and Xiong, Xiaobin and Ames, Aaron D and Goldman, Daniel I},
  booktitle={2018 IEEE-RAS 18th International Conference on Humanoid Robots (Humanoids)},
  pages={994--1001},
  year={2018},
  organization={IEEE}
}

@article{liu2025adaptive,
  title={Adaptive Locomotion on Mud through Proprioceptive Sensing of Substrate Properties},
  author={Liu, Shipeng and Tang, Jiaze and Meng, Siyuan and Qian, Feifei},
  journal={arXiv preprint arXiv:2504.19607},
  year={2025}
}

@article{lynch2020soft,
  title={The soft-landing problem: Minimizing energy loss by a legged robot impacting yielding terrain},
  author={Lynch, Daniel J and Lynch, Kevin M and Umbanhowar, Paul B},
  journal={IEEE Robotics and Automation Letters},
  volume={5},
  number={2},
  pages={3658--3665},
  year={2020},
  publisher={IEEE}
}

@article{roberts2021virtual,
  title={Virtual energy management for physical energy savings in a legged robot hopping on granular media},
  author={Roberts, Sonia F and Koditschek, Daniel E},
  journal={Frontiers in Robotics and AI},
  volume={8},
  pages={740927},
  year={2021},
  publisher={Frontiers Media SA}
}

@article{chang2020learning,
  title={Learning terrain dynamics: A gaussian process modeling and optimal control adaptation framework applied to robotic jumping},
  author={Chang, Alexander H and Hubicki, Christian M and Aguilar, Jeffrey J and Goldman, Daniel I and Ames, Aaron D and Vela, Patricio A},
  journal={IEEE Transactions on Control Systems Technology},
  volume={29},
  number={4},
  pages={1581--1596},
  year={2020},
  publisher={IEEE}
}

@article{wu2019tactile,
  title={Tactile sensing and terrain-based gait control for small legged robots},
  author={Wu, X Alice and Huh, Tae Myung and Sabin, Aaron and Suresh, Srinivasan A and Cutkosky, Mark R},
  journal={IEEE Transactions on Robotics},
  volume={36},
  number={1},
  pages={15--27},
  year={2019},
  publisher={IEEE}
}

@article{choi2023learning,
  title={Learning quadrupedal locomotion on deformable terrain},
  author={Choi, Suyoung and Ji, Gwanghyeon and Park, Jeongsoo and Kim, Hyeongjun and Mun, Juhyeok and Lee, Jeong Hyun and Hwangbo, Jemin},
  journal={Science Robotics},
  volume={8},
  number={74},
  pages={eade2256},
  year={2023},
  publisher={American Association for the Advancement of Science}
}

@inproceedings{chang2017learning,
  title={Learning to jump in granular media: Unifying optimal control synthesis with Gaussian process-based regression},
  author={Chang, Alexander H and Hubicki, Christian M and Aguilar, Jeff J and Goldman, Daniel I and Ames, Aaron D and Vela, Patricio A},
  booktitle={2017 IEEE International Conference on Robotics and Automation (ICRA)},
  pages={2154--2160},
  year={2017},
  organization={IEEE}
}

@inproceedings{hubicki2016tractable,
  title={Tractable terrain-aware motion planning on granular media: An impulsive jumping study},
  author={Hubicki, Christian M and Aguilar, Jeff J and Goldman, Daniel I and Ames, Aaron D},
  booktitle={2016 IEEE/RSJ International Conference on Intelligent Robots and Systems (IROS)},
  pages={3887--3892},
  year={2016},
  organization={IEEE}
}

@article{liao2025bio,
  title={A bio-inspired sand-rolling robot: effect of body shape on sand rolling performance},
  author={Liao, Xingjue and Liu, Wenhao and Wu, Hao and Qian, Feifei},
  journal={arXiv preprint arXiv:2503.13919},
  year={2025}
}

@article{li2023need,
  title={The need for and feasibility of alternative ground robots to traverse sandy and rocky extraterrestrial terrain},
  author={Li, Chen and Lewis, Kevin},
  journal={Advanced Intelligent Systems},
  volume={5},
  number={3},
  pages={2100195},
  year={2023},
  publisher={Wiley Online Library}
}

@article{li2013terradynamics,
  title={A terradynamics of legged locomotion on granular media},
  author={Li, Chen and Zhang, Tingnan and Goldman, Daniel I},
  journal={science},
  volume={339},
  number={6126},
  pages={1408--1412},
  year={2013},
  publisher={American Association for the Advancement of Science}
}

@inproceedings{kenneally2018actuator,
  title={Actuator transparency and the energetic cost of proprioception},
  author={Kenneally, Gavin and Chen, Wei-Hsi and Koditschek, Daniel E},
  booktitle={International Symposium on Experimental Robotics},
  pages={485--495},
  year={2018},
  organization={Springer}
}

@article{kenneally2016design,
  title={Design principles for a family of direct-drive legged robots},
  author={Kenneally, Gavin and De, Avik and Koditschek, Daniel E},
  journal={IEEE Robotics and Automation Letters},
  volume={1},
  number={2},
  pages={900--907},
  year={2016},
  publisher={IEEE}
}

@article{fulcher2025effect,
  title={Effect of Gait Design on Proprioceptive Sensing of Terrain Properties in a Quadrupedal Robot},
  author={Fulcher, Ethan and Caporale, J and Zhang, Yifeng and Ruck, John and Qian, Feifei},
  journal={arXiv preprint arXiv:2509.22065},
  year={2025}
}

@inproceedings{qian2013walking,
  title={Walking and running on yielding and fluidizing ground},
  author={Qian, Feifei and Zhang, Tingnan and Li, Chen and Masarati, Pierangelo and Hoover, Aaron M and Birkmeyer, Paul and Pullin, Andrew and Fearing, Ronald S and Goldman, Daniel I and Olin, FW},
  booktitle={Robotics: Science and Systems},
  pages={345},
  year={2013}
}

@article{godon2023maneuvering,
  title={Maneuvering on non-Newtonian fluidic terrain: a survey of animal and bio-inspired robot locomotion techniques on soft yielding grounds},
  author={Godon, Simon and Kruusmaa, Maarja and Ristolainen, Asko},
  journal={Frontiers in Robotics and AI},
  volume={10},
  pages={1113881},
  year={2023},
  publisher={Frontiers Media SA}
}

@article{aguilar2016robophysical,
  title={Robophysical study of jumping dynamics on granular media},
  author={Aguilar, Jeffrey and Goldman, Daniel I},
  journal={Nature Physics},
  volume={12},
  number={3},
  pages={278--283},
  year={2016},
  publisher={Nature Publishing Group UK London}
}

@article{lynch2024efficient,
  title={Efficient, Responsive, and Robust Hopping on Deformable Terrain},
  author={Lynch, Daniel J and Pusey, Jason L and Gart, Sean W and Umbanhowar, Paul B and Lynch, Kevin M},
  journal={IEEE Transactions on Robotics},
  year={2024},
  publisher={IEEE}
}

@article{goldman2008scaling,
  title={Scaling and dynamics of sphere and disk impact into granular media},
  author={Goldman, Daniel I and Umbanhowar, Paul},
  journal={Physical Review E—Statistical, Nonlinear, and Soft Matter Physics},
  volume={77},
  number={2},
  pages={021308},
  year={2008},
  publisher={APS}
}

@article{haddadin2017robot,
  title={Robot collisions: A survey on detection, isolation, and identification},
  author={Haddadin, Sami and De Luca, Alessandro and Albu-Sch{\"a}ffer, Alin},
  journal={IEEE Transactions on Robotics},
  volume={33},
  number={6},
  pages={1292--1312},
  year={2017},
  publisher={IEEE}
}

@article{morlando2021whole,
  title={Whole-body control with disturbance rejection through a momentum-based observer for quadruped robots},
  author={Morlando, Viviana and Teimoorzadeh, Ainoor and Ruggiero, Fabio},
  journal={Mechanism and Machine Theory},
  volume={164},
  pages={104412},
  year={2021},
  publisher={Elsevier}
}

@article{jin2019preparation,
  title={Preparation of sand beds using fluidization},
  author={Jin, Zhefei and Tang, Junyue and Umbanhowar, Paul B and Hambleton, James},
  year={2019},
  publisher={Engineering Archive}
}

@book{raibert1986legged,
  title={Legged robots that balance},
  author={Raibert, Marc H},
  year={1986},
  publisher={MIT press}
}

@article{kalman1960new,
  title        = {A New Approach to Linear Filtering and Prediction Problems},
  author       = {Kalman, R. E.},
  journal      = {Journal of Basic Engineering},
  volume       = {82},
  number       = {1},
  pages        = {35--45},
  year         = {1960},
  publisher    = {ASME},
  doi          = {10.1115/1.3662552}
}

@article{kang2018archimedes,
  title={Archimedes’ law explains penetration of solids into granular media},
  author={Kang, Wenting and Feng, Yajie and Liu, Caishan and Blumenfeld, Raphael},
  journal={Nature communications},
  volume={9},
  number={1},
  pages={1101},
  year={2018},
  publisher={Nature Publishing Group UK London}
}

@article{roth2021intrusion,
  title={Intrusion into granular media beyond the quasistatic regime},
  author={Roth, Leah K and Han, Endao and Jaeger, Heinrich M},
  journal={Physical Review Letters},
  volume={126},
  number={21},
  pages={218001},
  year={2021},
  publisher={APS}
}

@article{qian2019rapid,
  title={Rapid in situ characterization of soil erodibility with a field deployable robot},
  author={Qian, Feifei and Lee, Dylan and Nikolich, George and Koditschek, Daniel and Jerolmack, Douglas},
  journal={Journal of Geophysical Research: Earth Surface},
  volume={124},
  number={5},
  pages={1261--1280},
  year={2019},
  publisher={Wiley Online Library}
}

@article{ruck2024downslope,
  title={Downslope weakening of soil revealed by a rapid robotic rheometer},
  author={Ruck, John G and Wilson, Cristina G and Shipley, Thomas and Koditschek, Daniel and Qian, Feifei and Jerolmack, Douglas},
  journal={Geophysical Research Letters},
  volume={51},
  number={1},
  pages={e2023GL106468},
  year={2024},
  publisher={Wiley Online Library}
}

@article{liu2026scout,
  title={Scout-Rover cooperation: online terrain strength mapping and traversal risk estimation for planetary-analog explorations},
  author={Liu, Shipeng and Caporale, J and Zhang, Yifeng and Liao, Xingjue and Hoganson, William and Hu, Wilson and Misra, Shivangi and Peddinti, Neha and Holladay, Rachel and Fulcher, Ethan and others},
  journal={arXiv preprint arXiv:2602.18688},
  year={2026}
}

@article{hill2005scaling,
  title={Scaling vertical drag forces in granular media},
  author={Hill, Glen and Yeung, Susan and Koehler, Stephan A},
  journal={EPL (Europhysics Letters)},
  volume={72},
  number={1},
  pages={137--143},
  year={2005}
}

@article{gravish2014force,
  title={Force and flow at the onset of drag in plowed granular media},
  author={Gravish, Nick and Umbanhowar, Paul B and Goldman, Daniel I},
  journal={Physical Review E},
  volume={89},
  number={4},
  pages={042202},
  year={2014},
  publisher={APS}
}

@article{richefeu2006stress,
  title={Stress transmission in wet granular materials},
  author={Richefeu, Vincent and Radja{\i}, F and El Youssoufi, Moulay Sa{\i}d},
  journal={The european physical journal E},
  volume={21},
  number={4},
  pages={359--369},
  year={2006},
  publisher={Springer}
}

@article{marvi2014sidewinding,
  title={Sidewinding with minimal slip: Snake and robot ascent of sandy slopes},
  author={Marvi, Hamidreza and Gong, Chaohui and Gravish, Nick and Astley, Henry and Travers, Matthew and Hatton, Ross L and Mendelson III, Joseph R and Choset, Howie and Hu, David L and Goldman, Daniel I},
  journal={Science},
  volume={346},
  number={6206},
  pages={224--229},
  year={2014},
  publisher={American Association for the Advancement of Science}
}

@article{wang2022micromechanical,
  title={Micromechanical investigation of the particle size effect on the shear strength of uncrushable granular materials},
  author={Wang, Zi-Yi and Wang, Pei and Yin, Zhen-Yu and Wang, Rui},
  journal={Acta Geotechnica},
  volume={17},
  number={10},
  pages={4277--4296},
  year={2022},
  publisher={Springer}
}

%








\end{document}